\documentclass[pdflatex,sn-mathphys-num,iicol]{sn-jnl} 

\usepackage{graphicx}%
\usepackage{multirow}%
\usepackage{amsmath,amssymb,amsfonts}%
\usepackage{amsthm}%
\usepackage{mathrsfs}%
\usepackage[title]{appendix}%
\usepackage{xcolor}%
\usepackage{textcomp}%
\usepackage{manyfoot}%
\usepackage{booktabs}%
\usepackage{algorithm}%
\usepackage{algorithmicx}%
\usepackage{algpseudocode}%
\usepackage{listings}%
\usepackage{subcaption}  

\newcommand{\orcidlink}[1]{%
  \href{https://orcid.org/#1}{\orcidlogo}%
}

\theoremstyle{thmstyleone}%
\theoremstyle{thmstyletwo}%

\theoremstyle{thmstylethree}%

\begin{document}

\title[Article Title]{
AI-Augmented Pollen Recognition in Optical and Holographic Microscopy for Veterinary Imaging
}


\author*[1]{
\fnm{Swarn S.} \sur{Warshaneyan}\orcidlink{0000-0003-3815-6329}
}\email{swarn.warshaneyan@edi.lv}

\author[1]{\fnm{Maksims} \sur{Ivanovs}\orcidlink{0000-0003-2477-7327}}

\author[2]{\fnm{Blaž} \sur{Cugmas}\orcidlink{0000-0002-3615-7443}}

\author[2]{\fnm{Inese} \sur{Bērziņa}\orcidlink{0000-0002-9096-3542}}

\author[2]{\fnm{Laura} \sur{Goldberga} }

\author[2]{\fnm{Mindaugas} \sur{Tamosiunas}\orcidlink{0000-0001-5866-9557}}

\author[1]{\fnm{Roberts} \sur{Kadiķis}\orcidlink{0000-0001-6845-4381}}

\affil[1]{\orgname{Institute of Electronics and Computer Science}, \orgaddress{\street{14 Dzerbenes  Street}, \city{Riga}, \postcode{LV-1006}, \country{Latvia}}}
 
\affil[2]{\orgdiv{Faculty of Science and Technology}, \orgname{University of Latvia}, \orgaddress{\street{3 Jelgavas Street}, \city{Riga}, \postcode{LV-1004}, \country{Latvia}}}



\abstract{
We present a comprehensive study on fully automated pollen recognition across both conventional optical and digital in-line holographic microscopy (DIHM) images of sample slides. Visually recognizing pollen in unreconstructed holographic images remains challenging due to speckle noise, twin-image artifacts and substantial divergence from bright-field appearances. We establish the performance baseline by training YOLOv8s for object detection and MobileNetV3L for classification on a dual-modality dataset of automatically annotated optical and affinely aligned DIHM images. On optical data, detection mAP50 reaches 91.3\% and classification accuracy reaches 97\%, whereas on DIHM data, we achieve only 8.15\% for detection mAP50 and 50\% for classification accuracy. Expanding the bounding boxes of pollens in DIHM images over those acquired in aligned optical images achieves 13.3\% for detection mAP50 and 54\% for classification accuracy. To improve object detection in DIHM images, we employ a Wasserstein GAN with spectral normalization (WGAN-SN) to create synthetic DIHM images, yielding an FID score of 58.246. Mixing real-world and synthetic data at the 1.0 : 1.5 ratio for DIHM images improves object detection up to 15.4\%. These results demonstrate that GAN-based augmentation can reduce the performance divide, bringing fully automated DIHM workflows for veterinary imaging a small but important step closer to practice.
}




\maketitle

\section{INTRODUCTION}\label{sec1}

The usage of lens-less digital in-line holographic microscopy (DIHM) enables three-dimensional imaging without relying on external labeling dyes and with minimal sample preparation, presenting a lower cost and simpler alternative to conventional bright-field optical microscopy setups in veterinary imaging \cite{bib01}. Especially, when combined with automated object detection and classification techniques that rely on deep learning to minimize human intervention. However, typical DIHM unreconstructed images are plagued by speckle noise and twin-image artifacts with phase-contrast appearances that differ markedly from conventional microscopy images \cite{bib14}, which creates a significant domain gap for data-driven object recognition methods.

Convolutional neural networks have revolutionized biomedical image analysis by learning hierarchical features without needing manual selection \cite{bib04}. Architectures like U-Net can reliably segment cells and subcellular structures through encoder-decoder pathways \cite{bib04}. Lightweight classifiers like the MobileNet family achieve high accuracy under tight computational budgets \cite{bib06}. For object detection, the YOLO family unifies proposal and classification in a single pass \cite{bib07}.

Building on these advances, our recent conference work introduced a dual-modality pollen dataset comprising of manually annotated optical microscopy images and affinely aligned DIHM images \cite{bib08}. We transferred labels from optical to holographic images, followed by training YOLO v8s and MobileNetV3L as detection and classification backbones, respectively. With manually made object labels prepared by humans, the optical images yielded 46.2\% for mAP50 and 97\% for accuracy but the DIHM performance was 2.49\% for mAP50 and 42\% for accuracy. With an expanded dataset that used automatically made pollen grain labels prepared by a YOLO v8s model trained on a high-quality custom subset, the optical images yielded 91.3\% for mAP50 and 97\% for accuracy but the DIHM performance was 8.15\% for mAP50 and 50\% for accuracy. Afterwards, we expanded bounding box areas by up to 50\% to account for affine alignment noise, followed by training the detection model once again for DIHM images. We observed that the performance rose to 13.3\% for mAP50 and 54\% for accuracy as a consequence. 

Despite these improvements, this remains insufficient for practical deployment, which is especially true for detection results. Inspired by success in bridging modality gaps through augmentation of real-world data with synthetic data \cite{bib05}, we apply a Wasserstein GAN with spectral normalization (WGAN-SN) to create synthetic DIHM images. By trying out various ratios of the real-world and synthetic DIHM images as the training pool, we obtain detection performance of 15.4\% for mAP50 on holographic images with the best ratio.

The consequence of this work is a comprehensive baseline evaluation of YOLO v8s and MobileNetV3L on dual-modality pollen data \cite{bib08}, a WGAN-SN synthesis pipeline for DIHM image generation and systematic strategies for hybridizing real-world datasets with synthetic datasets that boost detection performance for DIHM images, serving as a precedent for similar use cases.

\section{RELATED WORK}\label{sec2}

Automated pollen recognition in bright-field microscopy has evolved from classical, hand-crafted feature approaches to end-to-end deep learning \cite{bib12}\cite{bib13}. Early systems used to extract morphology and texture descriptors, such as shape outlines and gray-level co-occurrence matrices, to train either support vector machines or random forests, which were able to achieve moderate accuracy under controlled illumination. The arrival of U-Net architectures allowed precise segmentation of cells and subcellular structures through symmetric encoder-decoder pathways \cite{bib04}. Transfer learning with pre-trained backbones (such as either ResNet or EfficientNet) further improved robustness across diverse specimen types and imaging conditions \cite{bib13}. Lightweight classifiers like MobileNetV3L, which was discovered by NAS, offer real-time inference with minimal loss in accuracy, making them well suited for either in-field or resource-constrained deployments \cite{bib06}.

The axial precision of DIHM was first demonstrated with living cells to subwavelength accuracy \cite{bib01} and subsequent work has applied deep networks to achieve image reconstruction and artifact suppression. DIHM reconstructions in the style of bright-field images from a single hologram have been achieved using cross-modality GANs, translating phase images into intensity counterparts with promising visual fidelity \cite{bib15}.

\begin{figure}[htbp]
  \centering
  \includegraphics[width=\columnwidth,keepaspectratio]{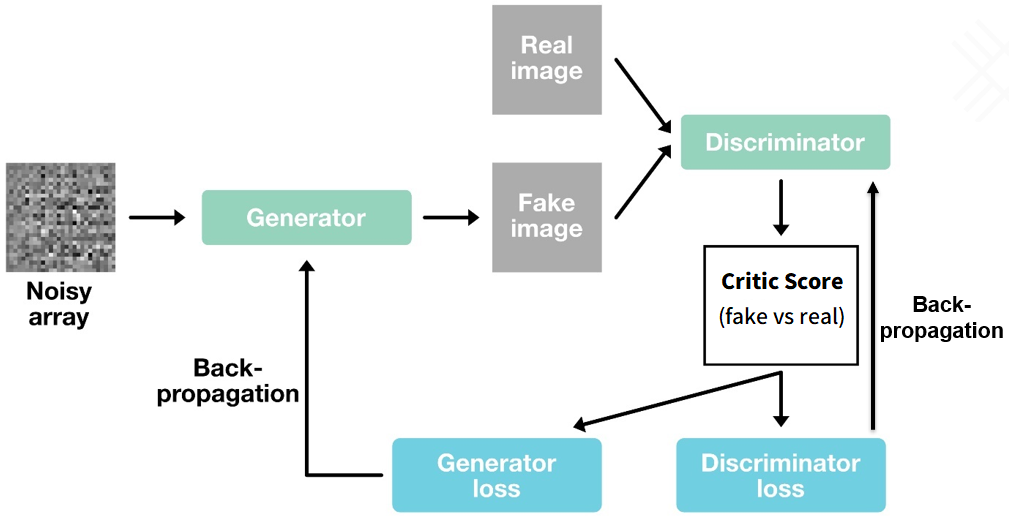}
  \caption{%
    A simplified visualization of the WGAN-SN architecture used in the project.%
  }
  \label{fig:wgan-design}
\end{figure}

Cross-modal translation and domain adaptation techniques have been proposed to mitigate modality gaps \cite{bib18}. A cycle-consistency network (a CycleGAN) learns bidirectional mappings between holographic and bright-field domains without paired data \cite{bib18}, while a physics-informed GAN incorporates optical forward models to constrain the image synthesis process \cite{bib19}. The use of domain randomization, which involves synthetic samples that are generated with wide parameter variations, has enabled robust transfer in robotics and vision tasks, suggesting potential for microscopy applications \cite{bib20}.

Generative data augmentation has proven effective in medical imaging contexts \cite{bib21}. The original GAN framework had introduced adversarial learning to achieve realistic image synthesis \cite{bib11}, which was followed by the Wasserstein GANs (the WGANs), that optimize an Earth-Mover distance for stable convergence \cite{bib09}. The technique of spectral normalization constraints the Lipschitz constant in the critic model, improving sample quality \cite{bib10}. In histopathology, the use of GAN-synthesized liver lesion images was able to boost liver-tumor classification performance by over 10\% \cite{bib21}. The utilization of diffusion probabilistic models offer an alternative high-fidelity synthesis pathway, although at greater computational cost.

Single-stage object detectors, such as the YOLO family, have been increasingly adopted for microscopy \cite{bib07}. YOLOv8s integrates dynamic anchor assignment and enhanced feature heads to improve detection of small and tightly packed objects \cite{bib07}, demonstrating superior inference speed when compared with two-stage methods. These detectors have been applied across biomedical domains, from blood-cell counting to subcellular particle tracking \cite{bib22}. Flow-cytometry in DIHM with deep learning has already been used to discriminate cell-death modalities \cite{bib16} and to enumerate circulating tumor cells in blood \cite{bib17}, demonstrating that direct object detection on raw holograms is feasible under controlled conditions. Still, their performance on holographic images remains underexplored.

Collectively, these strands of research (advanced classification and single-stage detection networks in microscopy imaging, combined with DIHM reconstruction and synthesis via GANs) form the foundation upon which our work builds. By generating high-quality synthetic holograms with WGAN-SN and systematically mixing them into detection training sets (Figure \ref{fig:wgan-design}), we aim to bridge the modality gap at least partially in order to enable reliable and automated pollen recognition using DIHM for veterinary imaging.

\section{DATA}\label{sec3}

\begin{figure}[htbp]
  \centering
  \includegraphics[width=0.5\columnwidth,keepaspectratio]{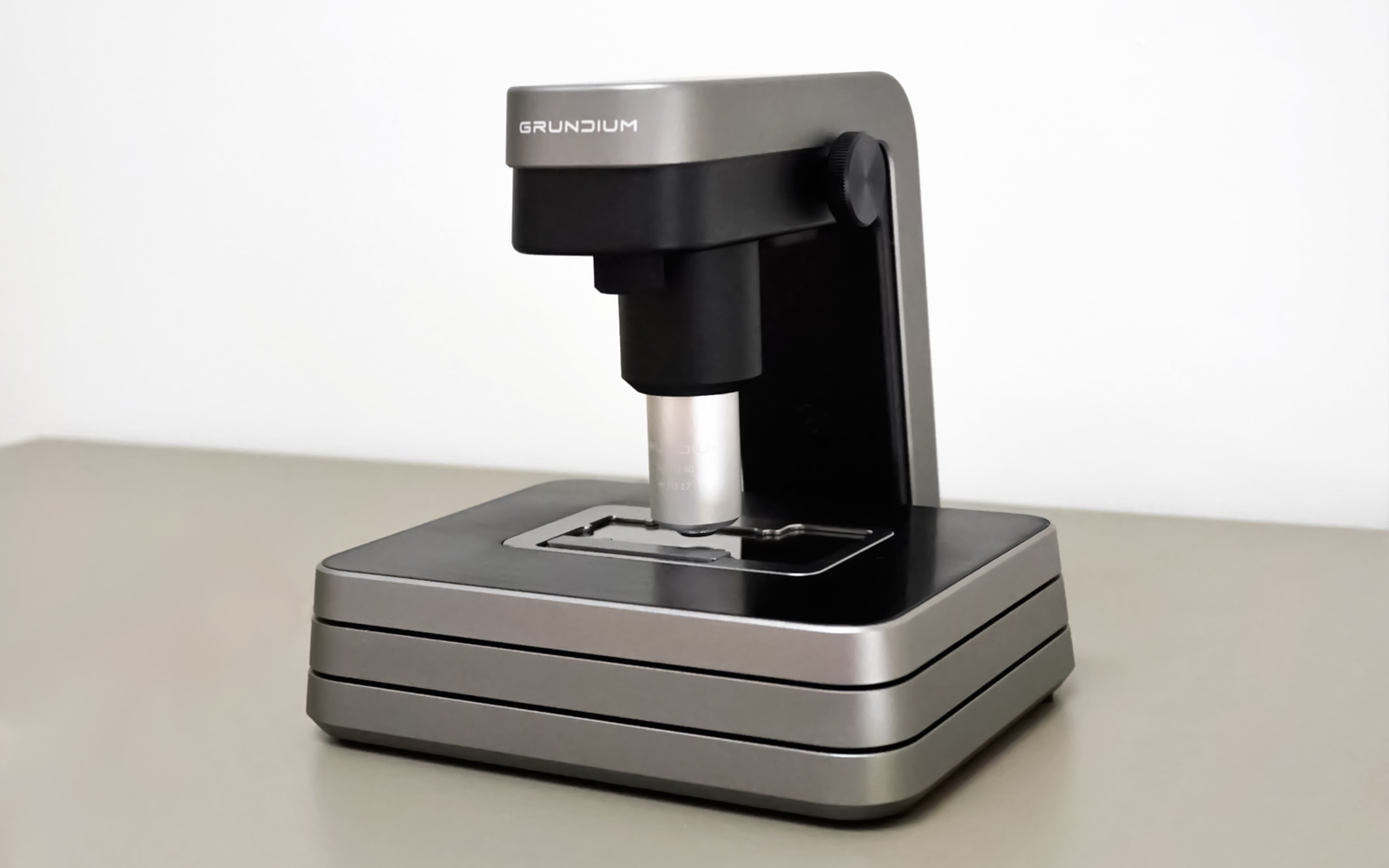}
  \caption{%
    The digital slide scanner Ocus 20, which was used for acquiring optical modality images.%
  }
  \label{fig:off-the-shelf-device}
\end{figure}

\begin{figure}[htbp]
  \centering
  \includegraphics[width=\columnwidth,keepaspectratio]{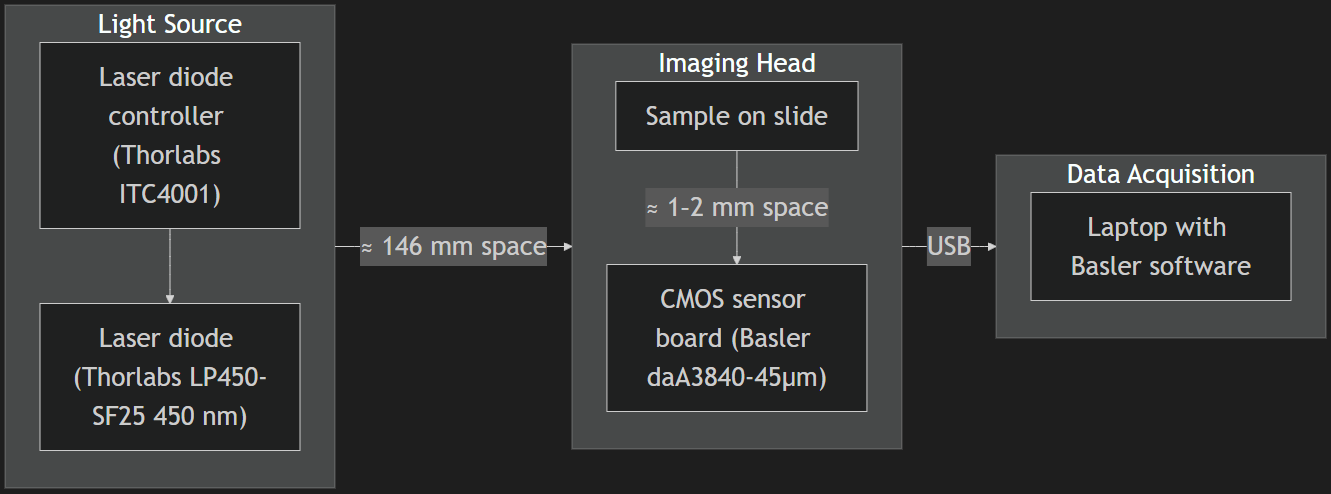}
  \caption{%
    A simplified outline of the custom greyscale holographic microscopy setup used for imaging. The design has only the essential components to minimize costs.%
  }
  \label{fig:custom-device}
\end{figure}

The object recognition model performance baseline creation step utilized real-world images of sample slides, containing pollen grains fixed through adhesive tape in a standard veterinary imaging manner, which were taken in two different imaging modalities from two separate microscopes: RGB format images from a brightfield optical microscope (Figure \ref{fig:off-the-shelf-device}) and greyscale format images from a custom-made lens-less holographic microscope (Figure \ref{fig:custom-device}). Although the effective number of manual labels for microscopy images of both modalities was different in the beginning, with all of 20 optical images and only 11 of 20 holographic images having manual labels, the sample size gap was successfully overcome by using automated labeling to give all available images bounding boxes, which made the difference practically negligible. For our current work, we focused on boosting detection model performance for holographic images. The real-world greyscale holographic images were used for training a WGAN-SN model for producing synthetic greyscale holographic images, with the intent of improving object detection performance for that imaging modality by augmenting the model training process with additional data.

\subsection{Real-World Samples}\label{subsec1}

We utilized 20 real-world brightfield optical images, with 20 real-world digital holographic images that were matched to the former due to their coverage of the same pollen grain sample slides (Table \ref{tab:real-overview}). The manually annotated bounding box labels for pollen grains were available for all of the real-world optical images. However, they were available for only 11 of the real-world holographic images. This hurdle was overcome by producing automated annotation labels using a YOLO v8s model trained on a high quality custom subset of the real-world optical images. All of the labels were transferred from the real-world optical images to the real-world holographic images using the process of affine aligning and this step allowed us to fill the sample size gap. We further reduced the performance gap using a bounding box area expansion strategy to make up for any discrepancies caused by imperfect label transfer in the affine alignment step.

\begin{table}[ht]
\caption{Real-World microscopy sample image dataset overview (without bounding box area expansion).}\label{tab:real-overview}%
\begin{tabular}{@{} 
    p{0.19\columnwidth}   
    p{0.13\columnwidth}   
    p{0.13\columnwidth}
    p{0.13\columnwidth}
    p{0.16\columnwidth}
  @{}}
\toprule

Image Type\newline (Modality) &
Hand-Labeled\newline Hi-Res Images &
Manual\newline Pollen Grain\newline Labels &
Auto-Labeled\newline Hi-Res Images & 
Automatic\newline Pollen Grain\newline Labels\\

\midrule

Optical (Raw) & 
20 & 
4536 & 
20 & 68268  \\
Holographic (Aligned) & 
11 & 
2437\footnotemark[1] & 
20 & 63018\footnotemark[1]  \\

\botrule
\end{tabular}
\footnotetext[1]{Some pollen grain labels fell outside the valid image area as a side-effect of the affine alignment process and were discarded.}
\end{table}

\subsection{Synthetic Samples}\label{subsec2}

\begin{figure}[htbp]               
  \centering
  \begin{subfigure}{0.48\columnwidth}
    \centering
    \includegraphics[width=\linewidth]{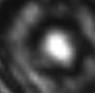}
    \caption{Real-World individual pollen grain image.}   
    \label{fig:real-individual}
  \end{subfigure}\hfill
  \begin{subfigure}{0.48\columnwidth}
    \centering
    \includegraphics[width=\linewidth]{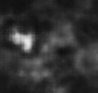}
    \caption{Synthetic individual pollen grain image.} 
    \label{fig:synthetic-individual}
  \end{subfigure}
  \vspace{1ex}                    
  \begin{subfigure}{0.48\columnwidth}
    \centering
    \includegraphics[width=\linewidth]{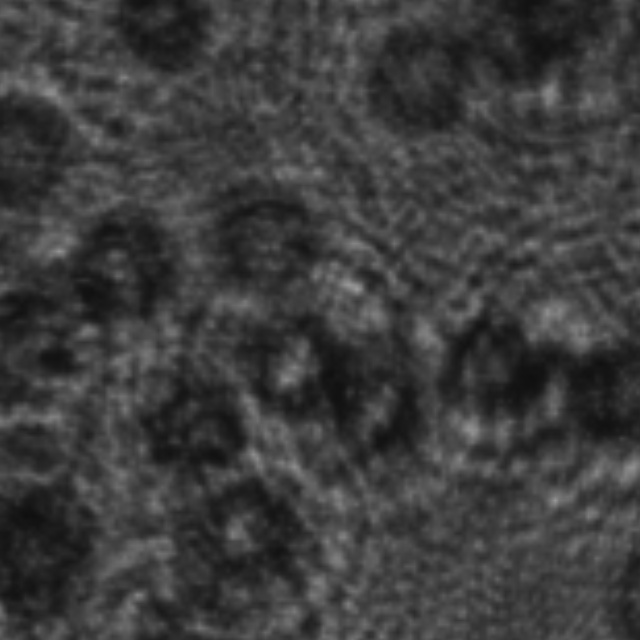}
    \caption{Real-World $640\times640$ sample image.}       
    \label{fig:real-sample}
  \end{subfigure}\hfill
  \begin{subfigure}{0.48\columnwidth}
    \centering
    \includegraphics[width=\linewidth]{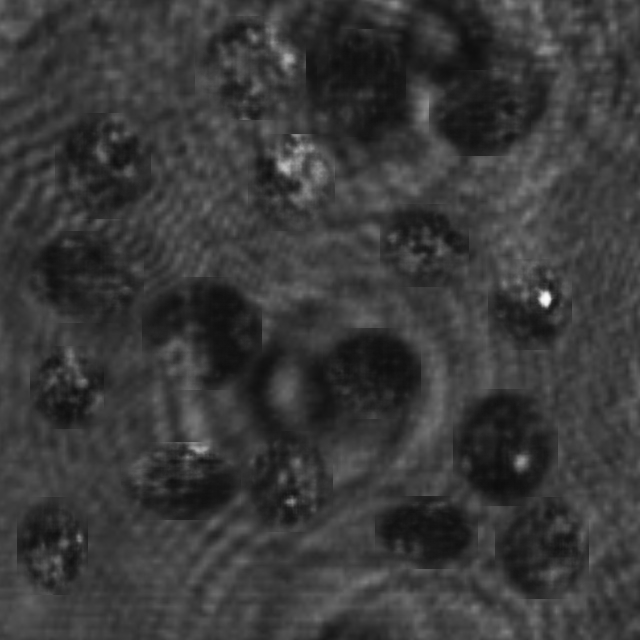}
    \caption{Composite $640\times640$ sample image.}  
    \label{fig:synthetic-sample}
  \end{subfigure}
  \caption{%
    Visual comparison of real-world and synthetic imagery used in the study.
    (a) and (b) show individual grains (real-world vs synthetic); (c) and (d) show
    full‑field $640\times640$ samples (real-world vs composite).%
  }
  \label{fig:real-versus-synthetic}
\end{figure}

Our approach to synthetic data creation begins by taking empty background patches without bounding boxes from the real-world images, followed by combining them with synthetic pollen grain images (Figure \ref{fig:real-versus-synthetic}) using alpha blending to create composite synthetic images (Figure \ref{fig:real-versus-synthetic}) for training the object detection model. We used a step size of 320 pixels to cut empty background patches of 640 by 640 pixels, getting a total of 5439 pieces. For comparison, we had a total of 4363 pieces of 640 by 640 pixels for the real-world image data set with 50\% bounding box expansion and automatically annotated labels. We had selected this specific version of the dataset from among the augmented versions because it had the best performance results in all model training steps involving real-world images. As the background patches are chosen randomly and get synthetic pollen grain images pasted on them randomly when creating composite images, we need to have more background patches than the number of composite images that we plan to create because this numerical difference allowed us to minimize chances of duplicates forming among the composite images. In our case, no duplication happened. 

We made use of the individual pollen grain images, with average dimensions of 94 $\pm$ 24 by 96 $\pm$ 25 pixels for training the chosen generative model. After separating the individual pollen grain images from the holographic microscopy sample slide images based on the bounding box labels, we cleaned the resulting data set by removing defective images, such as either removing blackened images caused due to dimensions mismatching between the optical and holographic images during the affine alignment step to avoid blackening artifacts appearing in synthetic images or removing lopsided images which have ratios worse than either 2:1 or 1:2 for height:width to ensure that the training images for the chosen generative model are as square-like as possible. 

For the dataset of automatically annotated bounding box labels of holographic images with 50\% area expansion to accommodate any alignment noise, we got 61671 instances from a total of 68268 instances after removing blackened images caused during the affine image aligning step and lopsided images caused during automated labeling step. We reduced outliers to improve the quality of the trained model. We used the cleaned dataset for training the generative model from scratch. In comparison, for the manually annotated bounding box labels, we got 2437 instances from a total of 2461 instances after removing the blackened images.

After the chosen generative model was trained, we created composite images for the next steps in two batches. We needed to create enough composite images for testing all of the planned data mixing ratios with minimal overlap, so we needed about 7635 composite images with about 109348 synthetic pollen grain images. The first batch had 62492 synthetic pollen grain images spread across 4363 composite images. The second batch had 46856 synthetic pollen grains spread across 3272 composite images.

\subsection{Mixing Real-World With Synthetic Samples}\label{subsec3}

 We had trained the object detection model with the real-world holographic image data set split in the train:validate:test configuration. We did not change either that or anything else, except the number of images in the training split for the training step's real-world data set, in order to ensure maximum reproducibility of results. The ratios were 700:15:15 with the counts as 3054 real-world items for training, 654 real-world items for validating and 655 real-world items for testing. We used the real-world:composite ratios of 1:1, 1:0.5, 1:1.5, 1:2 and 1:2.5 when mixing the images. All real-world and composite images used in the training step had dimensions of 640 by 640 pixels due to it being the ideal size for working with YOLO family models. The number of 640 by 640 images was used to calculate the ratios (Table \ref{tab:mixed-overview}). Each combination was used separately for training the object detection model and yielded different performance results. 
\begin{table}[ht]
\caption{An overview of the augmented model training image dataset.}\label{tab:mixed-overview}%
\begin{tabular}{@{} 
    p{0.22\columnwidth}   
    p{0.19\columnwidth}   
    p{0.19\columnwidth}
    p{0.19\columnwidth}
    p{0.19\columnwidth}
  @{}}
\toprule
Mixing Ratio For Training Split & Real-World $640\times640$ Pieces  & Composite $640\times640$ Pieces & Total $640\times640$ Pieces\\
\midrule
1:0\footnotemark[1]    & 3054\footnotemark[1]   & 0\footnotemark[1]  & 3054\footnotemark[1]\\
1:0.5    & 3054   & 1527  & 4581\\
1:1    & 3054   & 3054  & 6108\\
1:1.5    & 3054   & 4581  & 7635\\
1:2    & 3054   & 6108  & 9162\\
1:2.5    & 3054   & 7635  & 10689\\
\botrule
\end{tabular}
\footnotetext[1]{The baseline count with only real-world images is included for comparison purposes.}%
\end{table}
\section{EXPERIMENTS}\label{sec4}

Our choice of deep learning models was based on which models were state-of-the-art and can also work on edge devices without the need for significant changes in order to facilitate future deployment. We chose the YOLO v8s model because it has been designed to perform detection in a single forward pass with very low latency. We chose the MobileNetV3L model because it was designed to be lightweight, efficient, have very few parameters and also balance accuracy with latency. Among the many generative architectures, we chose the WGAN-SN as a pragmatic middle ground. Spectral normalisation adds almost no computational overhead while improving sample quality \cite{bib10}. The diffusion models and larger GAN variants (e.g. StyleGAN-v2) can achieve better FID, but at the cost of hundreds of sampling steps or tens of millions of parameters \cite{bib23}. The classical VAEs remain the lightest option but often produce perceptibly blurrier medical images \cite{bib24}. Recently improved versions of each family exist, yet the WGAN-SN met our edge-device budget without extra engineering.

\subsection{Establishing Performance Judgement Criteria}\label{subsec4}

We trained the chosen detection and classification models in separate steps with different versions of the data set to get a series of improving results. We applied only the automated labels for the optical images because there was no need for any bounding box area expansion based on the observed performance. We applied automated labels with 25\% and 50\% bounding box area expansion for the holographic images because there was a clear need for it based on the observed performance. We used the overall accuracy value as our main criteria for quality judgement with the MobileNetV3L model and the MAP50 value as our main criteria for quality judgement with the YOLO v8s model. The object detection for the holographic images seemed to require the most attention, causing us to focus on it during the next steps. 

\subsection{Generative Model Training}\label{subsec5}

We trained the chosen generative model in two consecutive phases with different learning rates, with all of the 61671 pollen grain images in the data pool to cover all of the diversity in the sample data set. Due to the fact that GAN training is unsupervised, label leakage is not a concern. Our main goal here was achieving a sufficient level of visual fidelity coverage to make the synthetic images realistic enough by providing a sufficient number of example images, rather than out-of-domain generalization, so not performing a split could not cause issues in this case, unlike when the generator model is expected to extrapolate beyond the given training distribution. We treated the Fréchet Inception Distance (FID) score primarily as a relative early-stopping signal for quality judgement. It was calculated after each training epoch using 10000 real-world and 10000 synthetic images. The training continued till the FID stopped decreasing (the ideal value is zero but achieving that is extremely rare). The ultimate criteria of generalization was the performance observed during downstream object detection. Any downstream performance gains provide practical evidence that the model did not simply memorize the training images. 


\subsection{Augmented Detection Model Training}\label{subsec6}

We trained the detection model with combined data set made using five different ratios of real-world:synthetic images (1:0.5, 1:1, 1:1.5, 1:2, 1:2.5). All combinations had exactly 3054 real-world images with dimensions of 640 by 640 pixels and the synthetic images had the same dimensions (Table \ref{tab:mixed-overview}). The variance existed in the form of the number of synthetic images that were added to each combination.

\begin{figure*}[t]
  \centering
  \begin{subfigure}{0.49\textwidth}
    \includegraphics[width=\linewidth]{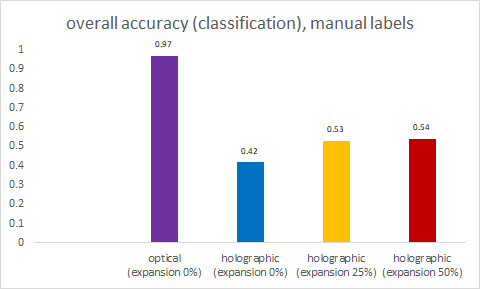}
    \caption{MobileNetV3L, manual labels}
    \label{fig:classification-manual}
  \end{subfigure}\hfill
  \begin{subfigure}{0.49\textwidth}
    \includegraphics[width=\linewidth]{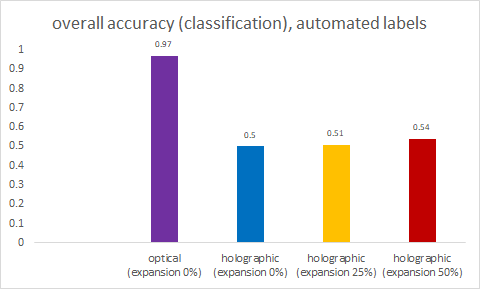}
    \caption{MobileNetV3L, automated labels}
    \label{fig:classification-automated}
  \end{subfigure}

  \vspace{1ex}

  \begin{subfigure}{0.49\textwidth}
    \includegraphics[width=\linewidth]{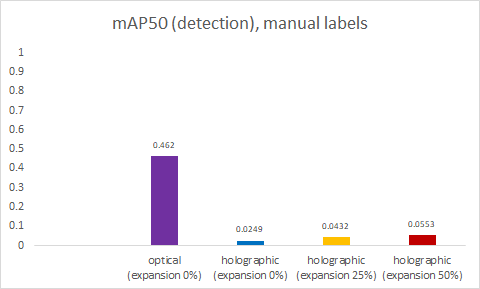}
    \caption{YOLO v8s, manual labels}
    \label{fig:detection-manual}
  \end{subfigure}\hfill
  \begin{subfigure}{0.49\textwidth}
    \includegraphics[width=\linewidth]{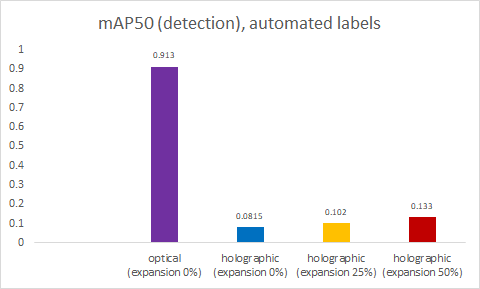}
    \caption{YOLO v8s, automated labels}
    \label{fig:detection-automated}
  \end{subfigure}

  \caption{Model performance at three bounding‑box scales on optical vs holographic images.}
  \label{fig:comparison-graphs}
\end{figure*}


\begin{figure}[htbp]
  \centering
  \includegraphics[width=0.9\columnwidth,keepaspectratio]{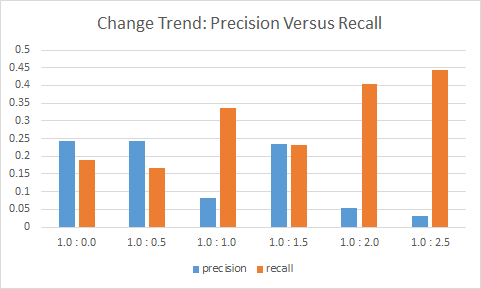}
  \caption{%
    A visualization of model performance change caused by mixing incrementally more synthetic data with real-world data.%
  }
  \label{fig:precision-vs-recall}
\end{figure}

\begin{figure}[htbp]
  \centering
  \includegraphics[width=0.9\columnwidth,keepaspectratio]{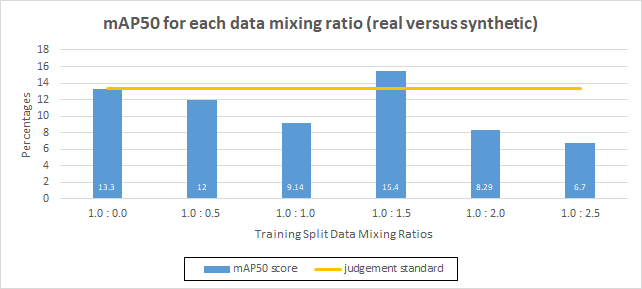}
  \caption{%
  A comparison of the mAP50 scores for various augmentation ratios (for mixing real and synthetic data) with the judgement standard.%
  }
  \label{fig:final-performance-comparison}
\end{figure}

\section{RESULTS}\label{sec5}

We can now go through the observations made during the experiments and compare them.

\subsection{Judging Real-World Image Performance}\label{subsec7}

We achieved 97\% overall accuracy with MobileNetV3L and 91.3\% mAP50 with YOLO v8s when using automated labels without any bounding box area expansion for optical images. In comparison, we achieved 97\% overall accuracy with MobileNetV3L and 46.2\% mAP50 with YOLO v8s when using manual labels without any bounding box area expansion. We achieved 54\% overall accuracy with MobileNetV3L and 13.3\% mAP50 with YOLO v8s when using automated labels with 50\% bounding box area expansion for holographic images. In comparison, we achieved 42\% overall accuracy with MobileNetV3L and 02.49\% mAP50 with YOLO v8s when using manual labels without any bounding box area expansion, which changed to 50\% overall accuracy and 08.15\% mAP50 upon switching to automated labels. The best results were for classification with automated labels and optical images, while the worst results were for detection with manual labels and holographic images (Figure \ref{fig:comparison-graphs}).

\subsection{Generating Synthetic Images}\label{subsec8}

Using the dataset version with automated labels to overcome sample size gaps and expanded bounding boxes to accommodate any alignment noise for training the WGAN-SN model, we got 68.32 as the FID score at epoch number 74 with a learning rate of 0.00005, after which we changed the learning rate to 0.00002, allowing us to get 65.4729 as the FID score at epoch number 82. We were unable to observe any further improvement for another 10 epochs, so we performed an additional step, where we filtered the synthetic images based on scores given by the critic (discriminator) network. We generated double the number of synthetic images than were needed before ranking them and discarding the bottom half. We used 60000 real-world and 60000 synthetic images this time for improving calculations, resulting in a slight FID score difference from the training step due to the significant difference in sample size. This gave 62.391 as the FID score before filtering and 58.246 as the FID score after filtering. After this, we moved to the next step, which involved using the synthetic images.

\subsection{Augmented Dataset Performance}\label{subsec9}

We mixed an incrementally increasing number of composite holographic images in a fixed number of real-world holographic images in the training split of the data set. For each combination of real-world and composite microscopy images, we retrained the detection model and compared it with the judgement standard, which was an mAP50 of 13.3\% for a 1:0 ratio, with a precision of 24.3\% and a recall of 19\%. The best performance obtained was an mAP50 of 15.4 for a 1:1.5 ratio, with a precision of 23.5\% and a recall of 23.2\%. This demonstrates an increase for mAP50 and recall with a decrease for precision.

\section{CONCLUSIONS}\label{sec6}

We observed noticeable changes downstream of our choices for the data preparation and experiments design steps. This confirmed that our decisions had impact. We observed a trend of increasing recall score and decreasing precision score (Figure \ref{fig:precision-vs-recall}) as more synthetic holographic images were mixed with the real-world holographic images in the training split of the data set. This caused the mAP50 value to have a downward trend as well, except for the 1.0:1.5 ratio, which had a very good balance of the recall and precision scores (Figure \ref{fig:final-performance-comparison}), which allowed it to outperform the judgement standard mAP50 value. The observations confirm that it is possible to achieve performance improvement by using synthetic data for augmenting real-world data. Since the similarity of synthetic images to real-world images directly affects what the object recognition models learn during the training phase, therefore model performance gains can be increased by lowering the FID score further to increase the overall visual quality of synthetic images. We recommend trying out synthetic image generation models offering relatively greater visual fidelity for achieving better performance improvements in any future work. 


\section*{DECLARATIONS}

\bmhead{\hspace{1em}Supplementary Information}
No supplementary files accompany this article.

\textbf{Project funding.}\quad
This research was funded by the Latvian Council of Science (project VetCyto, Nr. LZP-2023/1-0220).


\textbf{Competing interests.}\quad
The authors declare no competing interests.

\textbf{Ethics approval.}\quad
No ethics approval was required for this project.


\bibliography{sn-bibliography}

\end{document}